\pdfoutput=1

\documentclass[11pt]{article}

\usepackage{ACL2023}

\usepackage{times}
\usepackage{latexsym}

\usepackage[T1]{fontenc}

\usepackage[utf8]{inputenc}

\usepackage{microtype}

\usepackage{inconsolata}
\usepackage{booktabs}
\usepackage{multirow}
\usepackage{float}
\usepackage{adjustbox}
\usepackage{caption}
\usepackage{subcaption}
\usepackage{comment}
\usepackage{siunitx}
\usepackage{amsmath}
\newcommand{\dataset}{Echoes}
\newcommand{\datasetExp}{``Echoes from Alexandria''}
\newcommand{\wiki}{Echo-Wiki}
\newcommand{\wikiIntro}{Echo-XSum}
\newcommand{\extractive}{Echo-FairySum}
\newcommand{\numExtractive}{197}

\newcommand{\numWikiVersionsEN}{2,375}

\title{Echoes from Alexandria:\\A Large Resource for Multilingual Book Summarization}

 \author{Alessandro Scir\`e$^{1,2}$ \qquad Simone Conia$^{2}$ \qquad Simone Ciciliano$^{3}$\thanks{$^*$ Work carried out while at Sapienza University of Rome.} \qquad Roberto Navigli$^{2}$\\\\
\qquad Babelscape, Italy  \qquad \qquad ~~~~~
         Sapienza NLP Group \qquad ~~~~
         Free University of Bozen \\
  $^1$\texttt{scire@babelscape.com} \qquad  Sapienza University of Rome  \qquad $^3$\texttt{sciciliano@unibz.it} \\
         $^2$\texttt{\{first.lastname\}@uniroma1.it}}

\begin{document}
\maketitle
\begin{abstract}
In recent years, research in text summarization has mainly focused on the news domain, where texts are typically short and have strong layout features. 
The task of full-book summarization presents additional challenges which are hard to tackle with current resources, due to their limited size and availability in English only.
 To overcome these limitations, we present \datasetExp, or in shortened form, "\dataset{}", a large resource for multilingual book summarization.
 \dataset{} features three novel datasets: i) \wiki, for multilingual book summarization, ii) \wikiIntro, for extremely-compressive multilingual book summarization, and iii) \extractive, for extractive book summarization.
To the best of our knowledge, \dataset{} -- with its thousands of books and summaries -- is the largest resource, 
and the first to be multilingual, featuring 5 languages and 25 language pairs.
In addition to \dataset{}, we also introduce a new extractive-then-abstractive baseline, and, supported by our experimental results and manual analysis of the summaries generated, we argue that this baseline is more suitable for book summarization than purely-abstractive approaches.  
We release our resource and software at \url{https://github.com/Babelscape/echoes-from-alexandria} in the hope of fostering innovative research in multilingual book summarization.
\end{abstract}

\section{Introduction}
Recent research in Automatic Text Summarization -- the task of shortening a text while preserving its meaning -- has mainly focused on news stories. 
News texts are usually short documents; for example, 99.3\% and 98.6\% of the articles in XSum~\citep{XSum} and CNN/DailyMail~\citep{cnn-daily-mail}, respectively, are shorter than 2048 tokens.
Additionally, news stories are characterized by strong layout features, such as the ``lead bias'', in which the first sentences usually contain the most relevant information for a summary.
Accordingly, the Lead-3 baseline, which uses the first three sentences of a news item as its summary, performs competitively on news summarization benchmarks~\citep{gehrmann-etal-2018-bottom, lead3-XSum-cnndm}.
Although recent approaches have achieved high performance, it is still unclear how they behave on longer documents and whether they can generalize across domains and genres.
For this reason, the research community has been shifting toward more challenging settings, which include interviews~\citep{zhu-etal-2021-mediasum} and scientific articles~\citep{gupta-etal-2021-sumpubmed,cohan-etal-2018-discourse}.

One setting that has been attracting growing attention is full-book summarization~\citep{booksum}, i.e., the task of producing the plot of a book from its full text.
Summarizing a book is hard not only because of its average text length -- currently not processable in a single forward pass even by architectures for long-form text processing~\citep{longformer,guo-etal-2022-longt5} -- but also due to other critical aspects, such as the presence of dialogues, rich discourse structures, parallel and non-linear lines of plot, and long-distance dependencies between entities, among others.
Therefore, we deem book summarization a complex testbed to challenge current approaches and investigate their capabilities and limitations.

Although the first small-scale datasets for the task were introduced several years ago~\citep{book_summarization}, the area has recently regained traction thanks to larger-scale resources, such as BookSum~\citep{booksum} and NarrativeQA~\citep{narrativeqa}.
However, despite this recent progress, current resources for book summarization are still, i) limited in size, making them difficult to use for proper training and evaluation, and ii) monolingual (usually English-only).

To overcome these issues, we introduce \datasetExp~(\dataset), the largest resource to date for book summarization and the first one providing books and summaries in multiple languages.
We use \dataset{} to investigate how current summarization approaches perform on a large-scale multilingual summarization dataset, concluding that current purely-abstractive approaches still struggle in our setting.
We additionally devise a new baseline, showing that the extractive-then-abstractive paradigm represents a promising direction for future research. 

The main contributions of our work are the following:
\begin{itemize}
        \item We introduce \dataset, the first multilingual resource for book summarization, with thousands of texts and plots in 5 languages, for a total of 25 language pairs. \dataset~is also the largest resource among current English datasets for full-book summarization.
        
        \item We release the three datasets of \dataset{}: i) \wiki{}, for multilingual abstractive summarization, ii) \wikiIntro{}, for extremely-compressive multilingual book summarization, and iii) \extractive{}, an English dataset for evaluating extractive book summarization.
        
        \item We leverage BookSum and \dataset{} to evaluate state-of-the-art systems, both in zero-shot and fine-tuning settings, bringing to light their inadequate generalization capabilities in book summarization.
        
        \item Our experiments demonstrate that an \textit{extractive-then-abstractive} baseline outperforms the purely-abstractive counterpart on our datasets while achieving state-of-the-art results on BookSum. 
        
        \item We provide a comprehensive manual evaluation of the automatically generated summaries and release the dataset with our human judgments.
\end{itemize}
We hope our work will foster research in multilingual long document understanding and summarization.
We release \dataset~and our software for research purposes at \url{https://github.com/Babelscape/echoes-from-alexandria}.
\section{Related Work}

\paragraph{Resources for summarization.}
Research efforts to create summarization resources have steadily increased in numbers over recent years.
For the news domain, XSum~\citep{XSum} and CNN/DailyMail~\citep{cnn-daily-mail} are the \textit{de-facto} standard datasets for training and evaluating summarization systems.
XSum comprises 226k news articles accompanied by a one-sentence abstractive summary.
In CNN/DailyMail, the authors retrieved 93k articles from CNN\footnote{\href{https://www.edition.cnn.com/}{https://www.edition.cnn.com/}} and 220k articles from DailyMail\footnote{\href{https://www.dailymail.co.uk/}{https://www.dailymail.co.uk/}} newspapers.
Both publishers supplement their articles with a list of bullet points containing the main information of the news text.

More recently, summarization resources have been shifting towards more challenging scenarios, i.e., where the documents of interest are longer and belong to different domains.
Notably, \citet{cohan-etal-2018-discourse} released two large-scale datasets of long and structured scientific papers obtained from arXiv\footnote{\href{https://arxiv.org/}{https://arxiv.org/}} and PubMed\footnote{\href{https://pubmed.ncbi.nlm.nih.gov/}{https://pubmed.ncbi.nlm.nih.gov/}}.
In these datasets, paper abstracts are used as ground truth summaries.
Another relevant example is MediaSum~\citep{zhu-etal-2021-mediasum}, a collection of interview transcriptions from National Public Radio (NPR)\footnote{\href{https://www.npr.org/}{https://www.npr.org/}} and CNN, where overview and topic descriptions are employed as summaries.

In long-form text summarization research, a task that is attracting growing attention is book summarization.
Although this task was originally introduced several years ago by~\citet{book_summarization}, who released the first small-scale evaluation resource, book summarization regained traction thanks to a few notable endeavors. 
The most important example is BookSum~\citep{booksum}, which provides a collection of resources for book summarization at three levels of granularity: paragraph, chapter, and full book.
Book texts are collected from Project Gutenberg, while summaries are obtained from the Web Archive.\footnote{\href{https://web.archive.org/}{https://web.archive.org/}}
BookSum features 222 unique book titles with a total of 6,987 book chapters and 142,753 paragraphs.
Relatedly, NarrativeQA~\citep{narrativeqa} is a collection of 1572 stories retrieved from Project Gutenberg (783 books and 789 movie scripts) associated with summaries from Wikipedia. The annotators were required to generate questions and answers based on the summaries.
Even if NarrativeQA is primarily intended for Question Answering, it can also be used for book summarization. 
Due to their limited size, however, BookSum (in the full-book setting) and NarrativeQA can be more useful for evaluating models on the task rather than for training purposes. It is also worth noting that these resources are monolingual, i.e., English-only, limiting their usefulness for researchers seeking to evaluate multilingual summarization models.
Despite the great work carried out so far, we argue that there is still ample room to improve book summarization resources.

\paragraph{Approaches to book summarization.}
\citet{booksum} conducted experiments on full-book summarization using a generate\&rank strategy.
This approach involves training a system to generate paragraph-level summaries, which are then sorted by perplexity and concatenated to form a full-book summary.
More recently, \citet{openai-feedback-summ} proposed an approach where passages are recursively summarized and concatenated to form a full summary.
However, generated summaries are affected by the errors accumulated from previous stages~\citep{openai-feedback-summ}.
Recursively generating a summary is a paradigm that has also been used by other works for long-document summarization~\citep{summn, divide-and-conquer}.
Another family of approaches is that of \textit{extractive-then-abstractive} approaches.
This family of approaches first extracts key sentences from the input document and then uses such sentences as input to an abstractive model, which is tasked with generating a summary that captures the main ideas and themes of the source.
While it was successfully employed in previous works for short~\cite{li-etal-2021-ease} and long-form summarization~\cite{select-reinforce}, this paradigm has never been explored for summarizing books.
In this paper, we aim to fill this gap by presenting a new, simple extractive-then-abstractive model and showing its effectiveness for book summarization.

\begin{table*}[h]
\centering
\begin{adjustbox}{max width=\linewidth}
\begin{tabular}{l l c c c c c c}

\toprule

\multirow{2.5}{*}{\textbf{Dataset}} & \multirow{2.5}{*}{\textbf{Languages}} & \multirow{2.5}{*}{\textbf{\# Documents} {}} & \multirow{2.5}{*}{\textbf{Coverage}} & \multirow{2.5}{*}{\textbf{Density}} & \multirow{2.5}{*}{\textbf{C. Ratio}} & \multicolumn{2}{c}{\textbf{Avg. length (\# Tokens)}} \\

\cmidrule{7-8}

& & & & & & \textbf{Source} & \textbf{Summary} \\

\midrule

XSum & EN & 226,677 & 0.66 & 1.09 & ~~~~19.3 & ~~~~~~~438.4 & ~~~~~23.9 \\

CNN/DailyMail & EN & 311,971 & 0.85 & 3.47 & ~~~~14.9 & ~~~~~~~803.7 & ~~~~~59.7 \\

ArXiv/PubMed & EN & 346,187 & 0.87 & 3.94 & ~~~~31.2 & ~~~~5,179.2 & ~~~257.4 \\

MediaSum & EN & 463,596 & 0.80 & 1.86 & ~~116.3 & ~~~~1,925.8 & ~~~~~16.6\\

BookSum (full) & EN & ~~~~~~~405 
& 
0.89 & 1.83 & ~~126.2 & 112,885.2 & 1,167.2 \\

\midrule

    \wiki & EN, FR, DE, ES, IT & ~~~~5,001 
& 0.79 & 2.08 & ~~103.7 & ~~75,600.9 & ~~~729.4 \\
\wiki${_{EN}}$ & EN & ~~~~\numWikiVersionsEN
& 0.84 & 2.24 & ~~117.1 & ~~83,724.1 & ~~~678.0 \\

\wikiIntro & EN, FR, DE, ES, IT & ~~~~3,383 
& 0.78 & 1.67 & 1624.0 & ~~86,040.0 & ~~~~~53.0 \\
\wikiIntro${_{EN}}$ & EN & ~~~~1,828 
& 0.81 & 1.78 & 1706.1 & ~~90,971.9 & ~~~~~53.0 \\

\extractive & EN & ~~~~~~~197 
& 1.00 & 1.00 & ~~~~~~2.8 & ~~~~4,438.8 & 1,506.2 \\

\bottomrule

\end{tabular}
\end{adjustbox}

\caption{Comparison of \dataset{} (\wiki{}, \wikiIntro{}, and \extractive{}) with existing resources for summarization. \textbf{Coverage and density:} measures of the ``extractiveness'' of a summary. \textbf{Compression Ratio:} micro-average ratio between the lengths of the source and the summary.}

\label{tab:dataset-comparison}
\end{table*}

\section{\dataset} 
\dataset~is the first collection of resources for book summarization in 5 languages: English, French, German, Italian, and Spanish.
With \dataset{}, we introduce the following three novel datasets:
\begin{itemize}
    \item \textbf{\wiki{}}, in which we pair book texts with plots retrieved from a hand-curated list of Wikipedia page sections.
    
    \item \textbf{\wikiIntro{}}, in which we pair book texts with extremely-compressive summaries, manually created starting from the lead section of Wikipedia pages.

    \item \textbf{\extractive{}}, an evaluation dataset for extractive summarization of short stories and fairy tales, composed of \numExtractive~English manually-annotated extractive summaries. 
\end{itemize}
We provide an overview of the main differences between \dataset{} and existing resources in Table~\ref{tab:dataset-comparison}.

\subsection{Text collection}
\label{subsec:text-collection}
We collect the book texts that comprise \dataset{} from two main sources: Project Gutenberg and Wikisource.
Project Gutenberg is a digital library that provides free access to public-domain books and features over 60k texts.
We collect all the available books from Project Gutenberg by following their robot-access policies.\footnote{\href{https://www.gutenberg.org/help/mirroring.html}{https://www.gutenberg.org/help/mirroring.html}}
While often considered one of the most reliable sources of copyright-free books, Project Gutenberg provides only very limited coverage of non-English books and non-English translations of English books.
This is one of the reasons why we also rely on Wikisource.
Part of the Wikimedia Foundation, Wikisource contains a huge number of texts from a wide range of domains, e.g., books, and legal and historical documents, in various languages.
Therefore, for \dataset, we rely on Wikisource in English, French, German, Spanish, and Italian to retrieve other book texts and expand the coverage of books already available from Project Gutenberg.\footnote{Wikisource dumps are freely available to download at \href{https://dumps.wikimedia.org/enwikisource/}{https://dumps.wikimedia.org/<l>wikisource/} where <l> $\in$ \{ EN, FR, DE, ES, IT\}. Last accessed: July 1, 2022.}
We call this set of full-text books $B$.
We note that Wikisource can also be used to expand \dataset~to other languages.
Given the limited amount of work in multilingual summarization, we focus on the five above high-resource languages.
We defer the expansion of \dataset~to future work.

While Project Gutenberg has already been used as a source of books in previous resources, such as BookSum and NarrativeQA, the use of Wikisource is what enables \dataset~to become the largest resource for book summarization in English and the first resource for multilingual book summarization.

\subsection{Pairing books with Wikipedia summaries}
\label{subsec:echo-wiki}
Book summaries from Wikipedia follow a standard set of guidelines\footnote{\url{https://en.wikipedia.org/wiki/Wikipedia:How\_to\_write\_a\_plot\_summary}} and are often of remarkable quality, as they are continuously refined over time by the Wikipedia community. 
Therefore, once we have collected our set of full-book texts (see Section~\ref{subsec:text-collection}), we iterate over the Wikipedia dumps\footnote{Wikipedia dumps are freely available to download at \href{https://dumps.wikimedia.org/enwiki/}{https://dumps.wikimedia.org/<l>wiki/} where <l> $\in$ \{ EN, FR, DE, ES, IT\}. Last accessed: July 1, 2022.} in English, French, German, Italian, and Spanish. 
Given our set $B$ of full-book texts, and $W$, the set of Wikipedia pages, our objective is to uniquely associate a book $b \in B$ to a page $w \in W$, such that $w$ is the Wikipedia page of book $b$.
We obtain a set of potential matches by finding Wikipedia pages whose contents contain a hyperlink to a book in $B$. 
To improve the accuracy of our mapping, we first apply a string distance metric\footnote{We used the Edit distance to retain only those pairs whose titles were highly similar, by setting a stringent threshold (0.2).} to compare the titles of the books and their associated Wikipedia pages. We then check if the lead section of the Wikipedia page in question mentions the surname of the author of the associated book. This additional step helps us further refine and ensure the validity of our associations.

After our matching process, we manually inspect the cases in which books are associated with multiple Wikipedia pages. We discover that the pages in excess refer to adaptations of the book in other mediums, such as movies and theatrical plays. To resolve this ambiguity, we utilize the mapping between Wikipedia pages and Wikidata nodes to obtain metadata about the medium, e.g., \textit{book, movie, play}, and retain only the Wikipedia page that corresponds to the book.

At this point, given the Wikipedia page content, our goal is to extract only the book summary and discard other information, such as the biography of the author, historical background, prizes and accolades, and critical reception, among others.
To achieve this, we employ native speakers to manually identify a list of section names that, in the different languages, only contain plot information, aiming for high precision rather than coverage. We use the content of these identified sections as summaries and provide our list of section names in Appendix~\ref{app:section_names} for reference.
We name the resulting set of (Wikipedia summary, full-text book) pairs \textbf{\wiki}.

We note that the average number of unique editors (220.6),  revisions (421.4), and year of creation (2008) of the Wikipedia pages we select for the \wiki dataset are large: this indicates that their book summaries have been curated over time and suggests that they are of high quality. Table~\ref{tab:dataset-comparison} shows how \wiki{} compares against BookSum, the previous largest existing dataset for book summarization, to the best of our knowledge.
Besides being multilingual, it is worth noticing that \wiki{} is about 12 times larger than BookSum (5,001 vs. 405 books) while still featuring similar compression ratios (103.7 vs. 126.2).

\subsection{Enabling extreme summarization of books}
\label{subsec:echo-xsum}
Inspired by the work of \citet{XSum} on the news domain with XSum, which showcases the capabilities of highly-abstractive summarization, we introduce \textbf{\wikiIntro{}}, a new dataset for training and evaluating systems for extreme summarization of books. 
In \wikiIntro{}, we pair full-text books with very short summaries.
These summaries contain the minimum number of sentences required to provide an overview of the main contents of a book, typically one to three sentences.
The main challenge posed by \wikiIntro{} is dealing with the great disparity between the size of the input and the size of the output.
Indeed, as we can observe in Table~\ref{tab:dataset-comparison}, the compression ratio of \wikiIntro{} (1624.0) is unprecedented in the field of summarization, being an order of magnitude greater than those of \wiki{} (103.7) and BookSum (126.2).

The extreme summaries in \wikiIntro{} are the result of a manual annotation process, which involved an expert linguist who is a fluent speaker in all 5 languages of \dataset{}. The annotator was explicitly contracted for this task. Given a book and its previously-identified Wikipedia page (see Section~\ref{subsec:text-collection}), the annotator was tasked with extracting portions of text from the introduction that described the essential plot of a book.
An excerpt of a book text with the corresponding multilingual summaries from \wikiIntro{} can be found in Appendix \ref{app:echoxsum_excerpt}.
Notice that the portions of text extracted by the annotator are not necessarily contiguous, as long as the extracted text can be read independently of its original text.
As a rule of thumb for the annotation process, the linguist followed the definitions of Consistency, Relevance, Fluency, and Coherence of a summary \cite{fabbri-etal-2021-summeval}. The annotator spent an average of 5 minutes per sample.
We provide an example of the annotations produced in Appendix~\ref{app:echo_xsum_annotation}.
At the end of the manual creation of our extreme summaries, the resulting \wikiIntro{} is still about 8 times larger than BookSum (3,383 vs. 405 books).\footnote{\wikiIntro{} includes fewer book/summary pairs than \wiki{} because the annotator was not able to find an extreme summary in the Wikipedia pages of some books.}

\subsection{Classifying books into genres}
\label{subsec:book-classification}
Differently from existing resources, such as BookSum, which is limited by its relatively small size, the thousands of books in \dataset{} give us the opportunity to investigate book summarization more in-depth.
Indeed, books in \dataset{} cover a wide range of genres, including novels, theatrical plays, and poems, among others.
We argue that developing a strategy to automatically identify book genres provides valuable insights into the dataset and enables a fine-grained evaluation of current and future summarization approaches.
An analysis by genre can help us determine which genres are the most challenging to summarize.

\begin{figure}[t]
    \centering
    \includegraphics[width=1.0\linewidth]{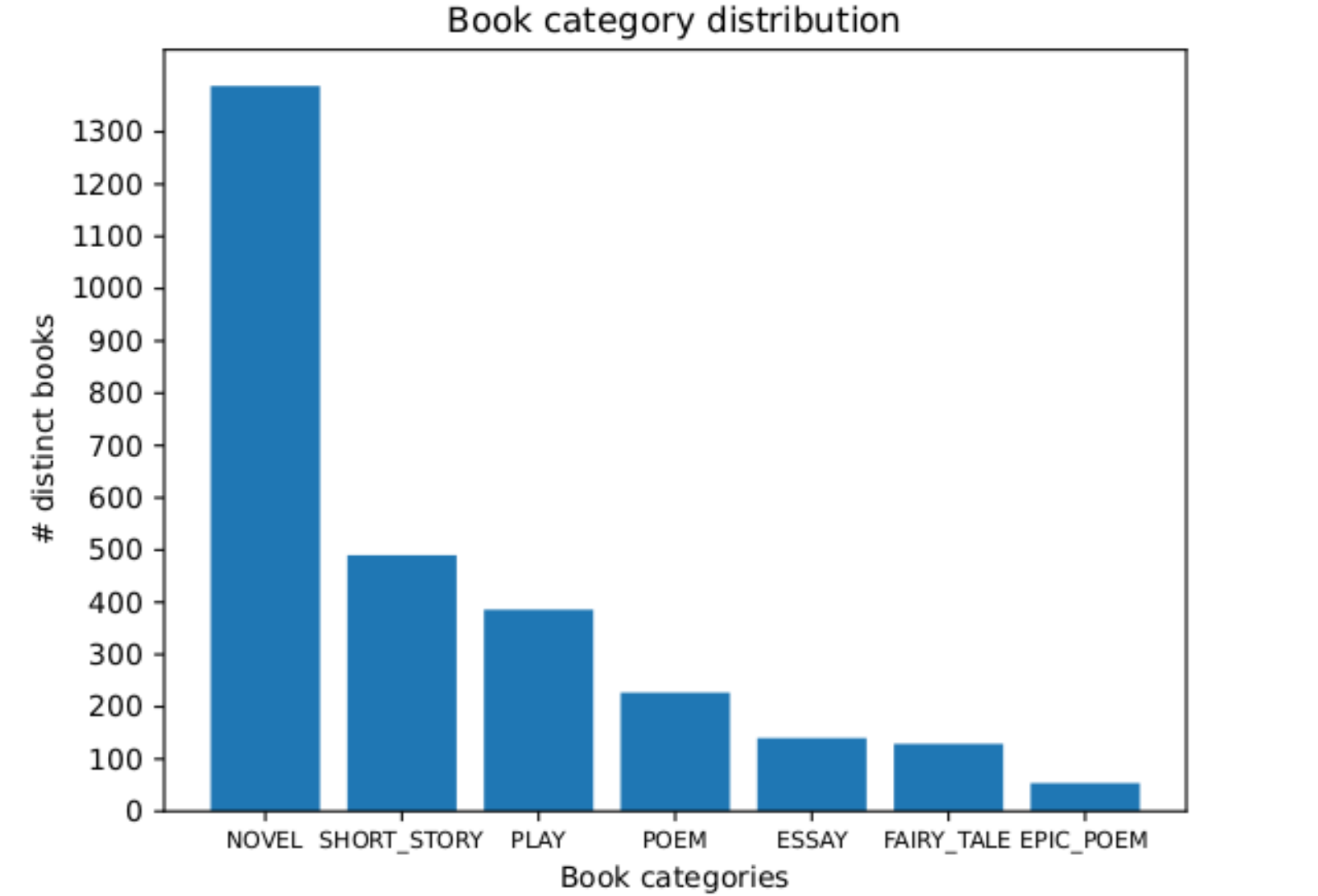}
    \vspace{-0.15cm}
    \caption{Distribution of the genres -- novels, short stories, play, poems, essays, fairy tales, and epic poems -- in the English partition of \wiki{}.}
    \label{fig:genre_distribution}
\end{figure}

Similarly to what was described in Section~\ref{subsec:echo-wiki}, we rely on a graph-based heuristic on the knowledge graph of Wikidata to identify genres.
More specifically, given a Wikipedia article of a book, we retrieve its corresponding Wikidata node, and analyze its relations (e.g., \textit{genre} and \textit{form\_of\_creative\_work}) with its neighboring nodes.
This process is able to distinguish between 7 main genres: novels, plays, poems, epic poems, short stories, fairy tales, and essays.
Note that our heuristic may assign more than one genre to a single book.
Figure~\ref{fig:genre_distribution} illustrates the distribution of the genres in the English partition of \wiki{}, showing that novels are the most represented genre, followed by short stories and plays.

\begin{figure*}[ht!]
  \centering
  \begin{subfigure}{.49\textwidth}
    \centering
    \includegraphics[width=\textwidth] {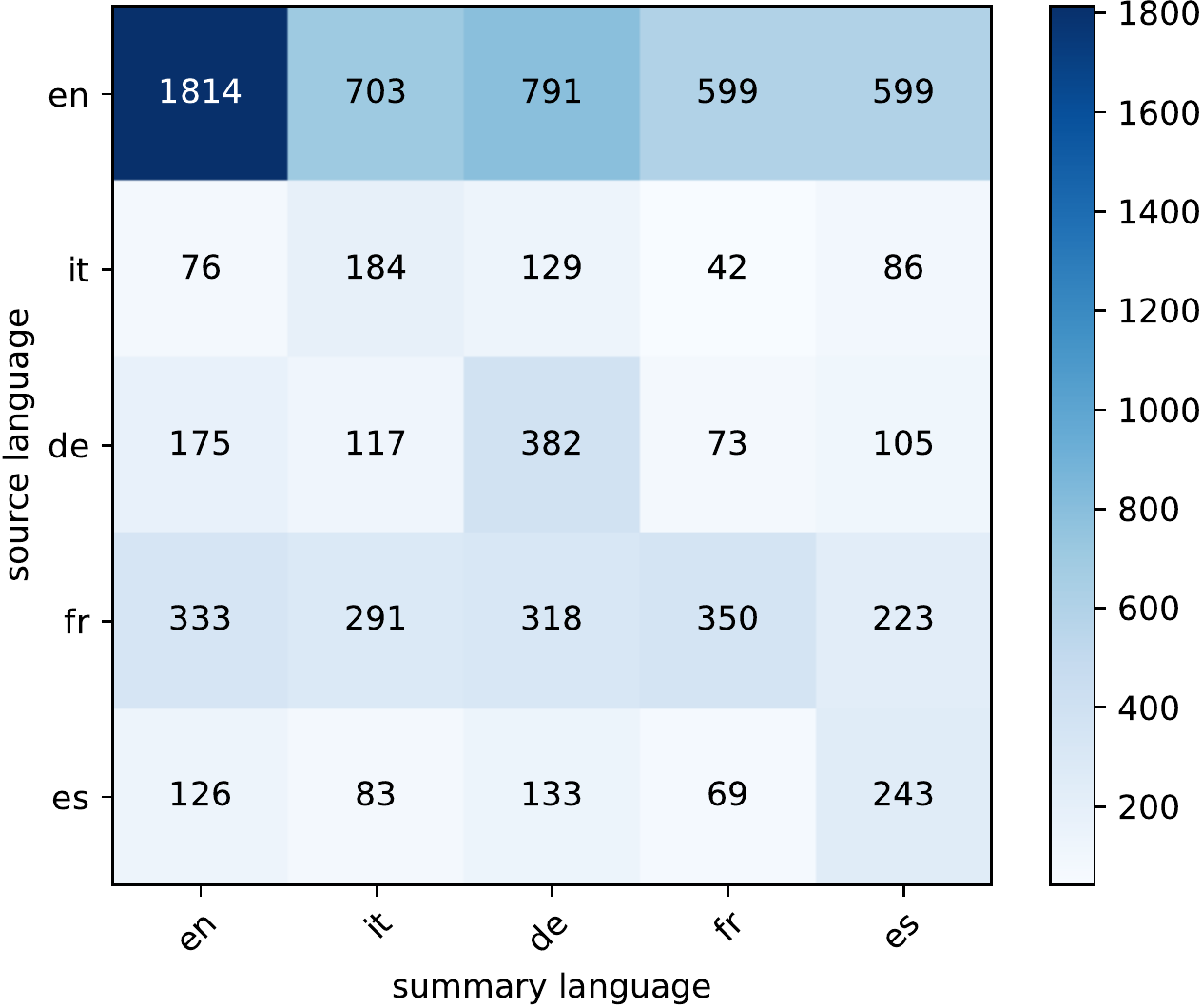}
    \label{subfig:distinct-cross-lingual}
  \end{subfigure}%
  \hspace{0.0025\textwidth}
  \begin{subfigure}{.49\textwidth}
    \centering
    \includegraphics[width=\textwidth]{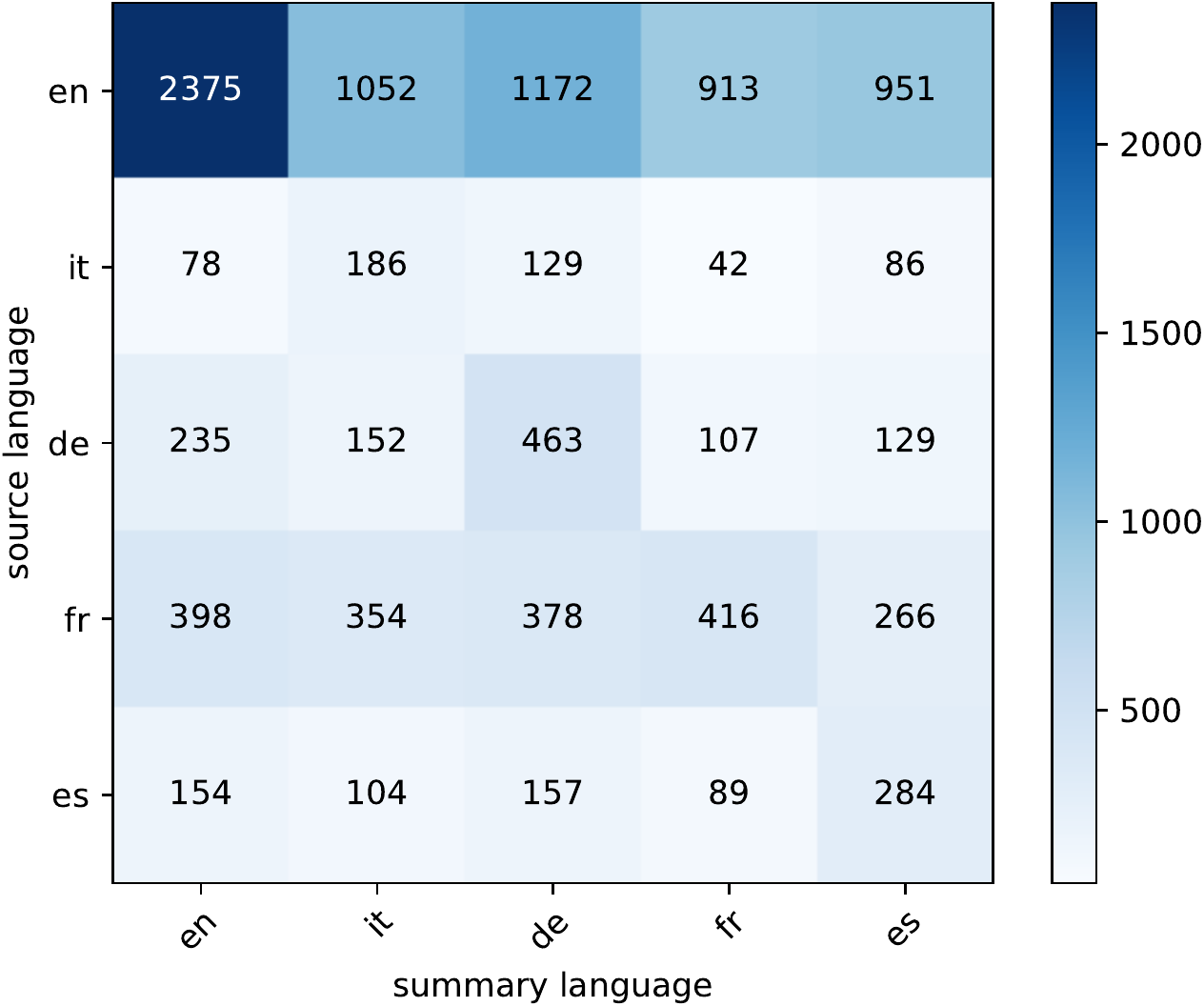}
    \label{subfig:all-cross-lingual}
  \end{subfigure}
\caption{Number of \textit{book-summary} (left) and \textit{version-summary} pairs (right) for all language pairs in \wiki{}. Best seen in color.}
\label{fig:cross-lingual}
\end{figure*}

\subsection{Digging up extractive summarization}
\label{subsec:echo-fairysum}
Over the past few years, the attention of the research community has gradually shifted from extractive to abstractive summarization, especially thanks to the advent of flexible sequence-to-sequence models, which have proven effective for summarizing short documents.
Thanks to genre classification (see Section~\ref{subsec:book-classification}), we are able to perform a small-scale investigation of extractive book summarization on two genres in \dataset{}.
More specifically, we construct \textbf{\extractive{}}, the first evaluation dataset for extractive summarization of fairy tales and short stories.

To create extractive summaries for \extractive{}, we set up the following manual annotation process: given the text of a book, and its abstractive summary from Wikipedia (Section \ref{subsec:echo-wiki}), annotators are required to extract relevant sentences from the book text.
A sentence is relevant if it provides a piece of information that is also contained in the abstractive summary.
The annotators were asked to adhere as closely as possible to the concepts of Consistency, Relevance, and Coherence defined by \citet{fabbri-etal-2021-summeval}.
The annotators were drawn from a pool of fifty-eight Master-level students from the `Narrative Understanding and Storytelling' minicourse held at the Sapienza University of Rome by the last co-author, as part of the AI and Robotics degree.
The selected students carried out the task as part of their course assignments.
On average, each student annotated 3 texts, resulting in multiple annotations for each text. The annotation agreement was measured using Cohen's Kappa coefficient, which indicated substantial agreement (0.71).
A subset of annotations was further validated by our contracted annotator to ensure that the students were adhering to the guidelines.
Overall, \extractive{} provides extractive summaries for 197 documents, 
about 4 times the size of the test set of BookSum.

\subsection{Aggregating books across versions and languages}
\label{subsec:book-aggregation}
A book can be published in various editions after its original publication.
Perhaps most importantly, the same version of a book can also be translated into multiple languages.
Given the potentially large variety of versions and translations of a book, we argue that it is important to aggregate those versions.
Indeed, aggregating books across versions and translations can allow \dataset{} to also be employed for machine translation, cross-lingual sentence alignment, and cross-lingual summarization.

To achieve this objective, we leverage two characteristics of Wikipedia.
First, we aggregate all those book texts aligned to the same Wikipedia page (see Section~\ref{subsec:echo-wiki}).
We increase the accuracy of this step by taking into account the information found on some Wikisource pages, which list the editions available for some books.
Second, we navigate the Wikipedia interlanguage links, which connect pages that refer to the same concept/entity in different languages, to aggregate different translations and summaries (in different languages) of the same book.
Figure \ref{fig:cross-lingual} presents the number of \textit{book-summary} and the \textit{version-summary} pairs for all the language pairs in \wiki{} obtained after our aggregation process. 

\section{Experiments and Results}
In recent years, two promising paradigms have emerged from previous work on long-document summarization: \textit{recursive-abstractive} and \textit{extractive-then-abstractive}.
In this section, we evaluate and analyze their effectiveness on \dataset{}.

\subsection{Recursive-abstractive approaches}
Recursive-abstractive approaches consist in dividing the source document into smaller segments, referred to as chunks, and then using an abstractive summarization model to summarize each segment.
If the concatenated output summaries are still larger than a single chunk, the recursive-abstractive approach repeats the process by treating the concatenation as a new source document and summarizing it in the same way. The recursive process continues until the concatenated output summaries are short enough to be considered as the final summary, i.e., until their size is shorter than the maximum size of a single chunk.

\begin{table}[t]
\footnotesize
\centering
\begin{adjustbox}{max width=\linewidth}
\begin{tabular}{l l c c c c}

\toprule

& \textbf{Model} & \textbf{R-1} &\textbf{R-2} & \textbf{R-L} & \textbf{BERTScore} \\

\midrule

\multirow{6}{*}{\rotatebox[origin=c]{90}{\textit{recursive-abs.}}}& BART$_\textit{XSum}$ & 18.02 & 2.91 & 13.81 & 0.438 \\

& BART$_\textit{MediaSum}$ & 13.95 & 5.11 & 12.72 & 0.416 \\

& LED$_\textit{XSum}$ & 18.86 & 2.99 & 14.83 & 0.440 \\ 

& LED$_\textit{MediaSum}$ & 14.69 & 4.26 & 12.79 & 0.421 \\

 & LongT5$_\textit{XSum}$ & 14.53 & 2.31 & 12.05 & 0.413 \\

 & LongT5$_\textit{MediaSum}$ & 16.54 & 5.47 & 14.35 & 0.429 \\

\midrule

\multirow{6}{*}{\rotatebox[origin=c]{90}{\textit{extractive-abs.}}} & BART & 30.44 & 12.41 & 25.76 & 0.557 \\

 & BART$_\textit{XSum}$ & \textbf{30.78} & 13.44 & \textbf{26.73} & 0.558 \\

 & LED & 30.18 & 12.73 & 25.79 & 0.558 \\

 & LED$_\textit{XSum}$ & 30.22 & 13.05 & 26.28 & \textbf{0.560} \\

 & LongT5 & 30.05 & \textbf{13.52} & 26.02 & \textbf{0.560} \\

 & LongT5$_\textit{XSum}$ & 29.42 & 13.35 & 26.00 & 0.557 \\
\bottomrule
\end{tabular}
\end{adjustbox}
\caption{Automatic evaluation of recursive-abstractive and extractive-then-abstractive approaches on \wikiIntro.}
\vspace{-2mm}
\label{tab:automatic_evaluation_xsum}
\end{table}

\paragraph{Experimental setting.}
In its simplest form, a recursive-abstractive approach requires a model trained on a standard summarization dataset; this model is then employed recursively, as described above.
For our experiments, we consider three sequence-to-sequence Transformer-based models -- BART-large~\cite{lewis-etal-2020-bart}, LED-base~\cite{longformer}, and LongT5-base~\cite{guo-etal-2022-longt5} -- and train them on XSum (short documents, news) and MediaSum (long documents, interviews).
Then, we evaluate our trained models on the test set of \wikiIntro{},\footnote{We split \wiki{} and \wikiIntro{} into train/dev/test sets using the standard 80/10/10 split.} whose summaries feature an average length similar to that of the summaries in XSum and MediaSum but belong to a different genre (books).
For the evaluation, we adopt standard summarization metrics, such as ROUGE-1, ROUGE-2, ROUGE-L, and BERTScore~\cite{bertscore}.

\paragraph{Results.}
Table~\ref{tab:automatic_evaluation_xsum} (top) provides an overview of the results obtained by our recursive-abstractive baseline using different language models and trained on different summarization datasets.
Overall, we can observe that, independently of the language model and training dataset employed, the baseline does not achieve good results on \wikiIntro{}.
Indeed, the best configuration (LED$_\textit{XSum}$) obtains only 14.83 points in ROUGE-L on \wikiIntro{}.
By comparison, the same configuration achieves 30.24 points on XSum.
Therefore, i) \wikiIntro{} is empirically more challenging than XSum, ii) a simple recursive-abstractive approach is not sufficient to obtain acceptable results on \wikiIntro{}, and, iii) different pretrained language models and different summarization datasets (from different genres/domains) do not significantly affect the results of a recursive-abstractive approach on our book summarization dataset.


\subsection{Extractive-then-abstractive approaches}
Since recursive-abstractive approaches yield unsatisfying results on \wikiIntro{} (see Table~\ref{tab:automatic_evaluation_xsum}), we propose a simple, novel baseline based on the extractive-then-abstractive paradigm.
Our model is composed of two submodules: the \textit{extractor} extracts key sentences from the input text, while the \textit{abstractor} uses the concatenation of these key sentences to generate an abstractive plot of the book.
Given an input text $T = (s_1, s_2, \dots, s_{|T|})$ where each $s_i$ is a sentence, the extractor produces a score in $[0.0, 1.0]$ for each $s_i$, quantifying its degree of importance for the target summary.
More formally:
\begin{align*}
    \mathbf{e}_i^s &= \textsc{\small SentenceEncoder}(s_i) \\
    \textsc{\small Score}(s_i) &= \sigma(W\mathbf{e}_i + \mathbf{b})
\end{align*}
where $\mathbf{e}^s_i$ is the sentence representation of $s_i$ from a \textsc{\small SentenceEncoder}.\footnote{We adopt a SentenceTransformer based on Distil-RoBERTa from \href{https://www.sbert.net/}{https://www.sbert.net/}.}
Then, the abstractor takes the subset $T^*$ composed of the $k$ sentences with higher scores according to the extractor, and uses $T^*$ to generate the final summary.
To make the abstractor aware of the relative importance of each sentence, we multiply the embedding of each token by the score of its sentence, as follows:
\begin{align*}
    \mathbf{e}^t_{i,j} = \textsc{\small  Score}(s_i)\ \cdot\ \textsc{\small Embedding}(t_{i,j})
\end{align*}
where $\mathbf{e}^t_{i,j}$ is the encoding of the $j$-th token of the $i$-th sentence, for each sentence in $T^*$.

The model is trained in an end-to-end fashion, i.e., the extractor and abstractor are trained jointly, by minimizing the cross-entropy loss between the reference summary and the generated summary.

\subparagraph{Experimental setting.}
We follow the experimental setting we used for our recursive-abstractive approach.
We train and evaluate 3 models -- BART-large, LED-base, and LongT5-base -- on \wikiIntro{}.
Since pretraining on XSum results in slightly improved performance for the recursive-abstractive approach, we also evaluate how pretraining on XSum affects the performance of our extractive-then-abstractive approach.
Finally, we also train and evaluate our approach on \wiki{} and on BookSum (the latter to directly compare performance with the current state of the art).

\begin{table}[t]
\footnotesize
\centering

\begin{adjustbox}{max width=\linewidth}
\begin{tabular}{l l l c c c c}

\toprule

\textbf{Model} & \textbf{R-1} &\textbf{R-2} & \textbf{R-L} & \textbf{BERTScore} \\

\midrule

BART & 16.64 & 4.07 & 13.09 & 0.517 \\

LED & 19.13 & 4.89 & 14.74 & 0.532 \\

LongT5 & \textbf{27.20} & \textbf{6.87} & \textbf{19.74} & \textbf{0.561} \\

\bottomrule

\end{tabular}
\end{adjustbox}
\caption{Automatic evaluation of extractive-then-abstractive approaches on \wiki.}
\label{tab:automatic_evaluation_wiki}
\end{table}

\begin{table}[t]
\centering
\begin{adjustbox}{max width=\linewidth}
\begin{tabular}{l l c c c c}
\toprule
& \textbf{Model} & \textbf{Cons.} & \textbf{Fluency} & \textbf{Rel.} & \textbf{Coher.} \\
\midrule
\multirow{6}{*}{\rotatebox[origin=c]{90}{\textit{recursive-abs.}}} & BART$_\textit{XSum}$ & 2.19 & 3.81 & 1.62 & 3.58 \\
& LED$_\textit{XSum}$ & 1.65 & 3.96 & 1.31 & 2.92 \\
& LongT5$_\textit{XSum}$ & 1.23 & 2.88 & 1.19 & 2.34 \\
& BART$_\textit{MediaSum}$ & 1.73 & 2.46 & 1.62 & 2.19 \\
& LED$_\textit{MediaSum}$ & 1.61 & 2.23 & 1.46 & 1.92 \\
& LongT5$_\textit{MediaSum}$ & 1.11 & 1.38 & 1.12 & 1.38 \\
\midrule
\multirow{6}{*}{\rotatebox[origin=c]{90}{\textit{extractive-abs.}}} & BART & 1.69 & 4.38 & 1.76 & 4.42 \\
& BART$_\textit{XSum}$ & 1.61 & 3.06 & 1.35 & 2.71 \\
& LED & 1.84 & 4.34 & 1.84 & 4.23 \\
& LED$_\textit{XSum}$ & 1.72 & 3.97 & 1.55 & 3.66 \\
& LongT5 & \textbf{2.73} & \textbf{4.50} &\textbf{2.73} & \textbf{4.62} \\
& LongT5$_\textit{XSum}$ & 2.04 & 3.85 & 1.74 & 3.52 \\
\bottomrule
\end{tabular}
\end{adjustbox}
\caption{Human evaluation of recursive-abstractive approaches on \wikiIntro.}
\label{tab:human_evaluation_xsum}
\end{table}

\begin{table}[t]
\footnotesize
\centering
\begin{adjustbox}{max width=\linewidth}
\begin{tabular}{l l c c c c}
\toprule
& \textbf{Model} & \textbf{Cons.} & \textbf{Fluency} & \textbf{Rel.} & \textbf{Coher.} \\
\midrule
& BART & 2.06 & \textbf{3.73} & 1.65 & \textbf{3.08} \\
& LED & 2.02 & 3.63 & 1.61 & 3.07 \\
& LongT5 & \textbf{2.15} & 3.62 & \textbf{1.72} & 3.06 \\
\bottomrule

\end{tabular}
\end{adjustbox}
\caption{Human evaluation of extractive-then-abstractive approaches on \wiki.}
\label{tab:human_evaluation_wiki}
\end{table}

\begin{table*}[t]
\centering
\begin{tabular}{l c c c c}

\toprule

\textbf{Approach} & \textbf{R-1} & \textbf{R-2} & \textbf{R-L} & \textbf{\# Params.} \\

\midrule

\citet{booksum} & 39.87 & ~~8.01 & 13.99 & ~~~~~~~737M\\

\citet{openai-feedback-summ} & 43.19 & 10.63 & 17.10 & 175,000M\\

\textbf{Ours }(LED/extractive-abs.) & 42.13 & 10.53 & 16.75 & ~~~~~~~243M \\

\bottomrule
\end{tabular}

\caption{Results of our approach compared to the state of the art on the BookSum test set.}
\label{tab:booksum_results}

\end{table*}

\begin{table*}[h]
\centering
\begin{tabular}{ c c c c c c }
\toprule
\textbf{Language} & \textbf{\# Examples} & \textbf{R-1} & \textbf{R-2} & \textbf{R-L} & \textbf{BERTScore} \\
\midrule
de & 24 & 21.219 & 6.808 & 17.742 & 0.641 \\ 
fr & 33 & 21.602 & 7.681 & 17.721 & 0.622 \\ 
es & 45 & 24.509 & 8.966 & 19.554 & 0.634 \\ 
it & 37 & 25.174 & 10.446 & 22.343 & 0.633 \\
\bottomrule
\end{tabular}
\caption{\textit{Summarize-then-translate} experiment. We translate the summaries generated by LongT5$_{base}$ model, fine-tuned on \wikiIntro, and compare them against gold standard references.}
\label{tab:summarize-then-translate}
\vspace{-2mm}
\end{table*}

\paragraph{Results.}
Table~\ref{tab:automatic_evaluation_xsum} (bottom) provides an overview of the results obtained by our extractive-then-abstractive approach on \wikiIntro{}.
We can immediately notice that each configuration significantly outperforms the recursive-abstractive baselines by a large margin.
For example, the best extractive-then-abstractive model (BART$_\textit{XSum}$) improves over the best recursive-abstractive model (LED$_\textit{XSum}$) by 11.90 points in ROUGE-L (26.73 vs. 14.83), and this is true for all the metrics we consider (ROUGE-1, ROUGE-2, ROUGE-L, and BERTScore).
It is interesting to note that, while there is little difference in the results on \wikiIntro{} of different model configurations, there is a significant difference between BART, LED, and LongT5 when evaluated on \wiki{}, as shown in Table~\ref{tab:automatic_evaluation_wiki}.
We hypothesize that such a variance in performance is due to several factors, but the inadequacy of current non-semantic metrics plays a large role, as supported by our human evaluation (see Section~\ref{sec:analysis_and_discussion}).

Finally, we further assess the effectiveness of our extractive-then-abstractive approach on the standard test set of BookSum (Table \ref{tab:booksum_results}).
In particular, our approach outperforms the system of \citet{booksum} using 33\% of its parameters, and is competitive with the system of \citet{openai-feedback-summ} using only 0.1\% of its parameters.

\section{Analysis and Discussion}
\label{sec:analysis_and_discussion}
\paragraph{Human evaluation.}
Following common practice in the field of summarization, we set up a human evaluation process to assess the quality of the system-generated summaries.
The annotation task, performed by an expert English speaker, consists of reading the source text and rating the summaries using a Likert scale for Consistency, Relevance, Fluency, and Coherence, as outlined in \citet{fabbri-etal-2021-summeval}.
To make this experiment feasible in terms of time and resources, we focus our evaluation on fairy tales and short stories, which can be read by a human in a short time.
Interestingly, but not surprisingly~\cite{fabbri-etal-2021-summeval}, the results of our human evaluation experiment tell a story that is different from ROUGE, as shown in Tables~\ref{tab:human_evaluation_xsum} and \ref{tab:human_evaluation_wiki}.
However, the evaluation still highlights the effectiveness of our extractive-then-abstractive model compared to the recursive-abstractive baseline.
It is clear, however, that future work should focus in particular on improving the Consistency and Relevance of the summaries generated.
\begin{figure}[t]
    \centering
    \includegraphics[width=1.0\linewidth]{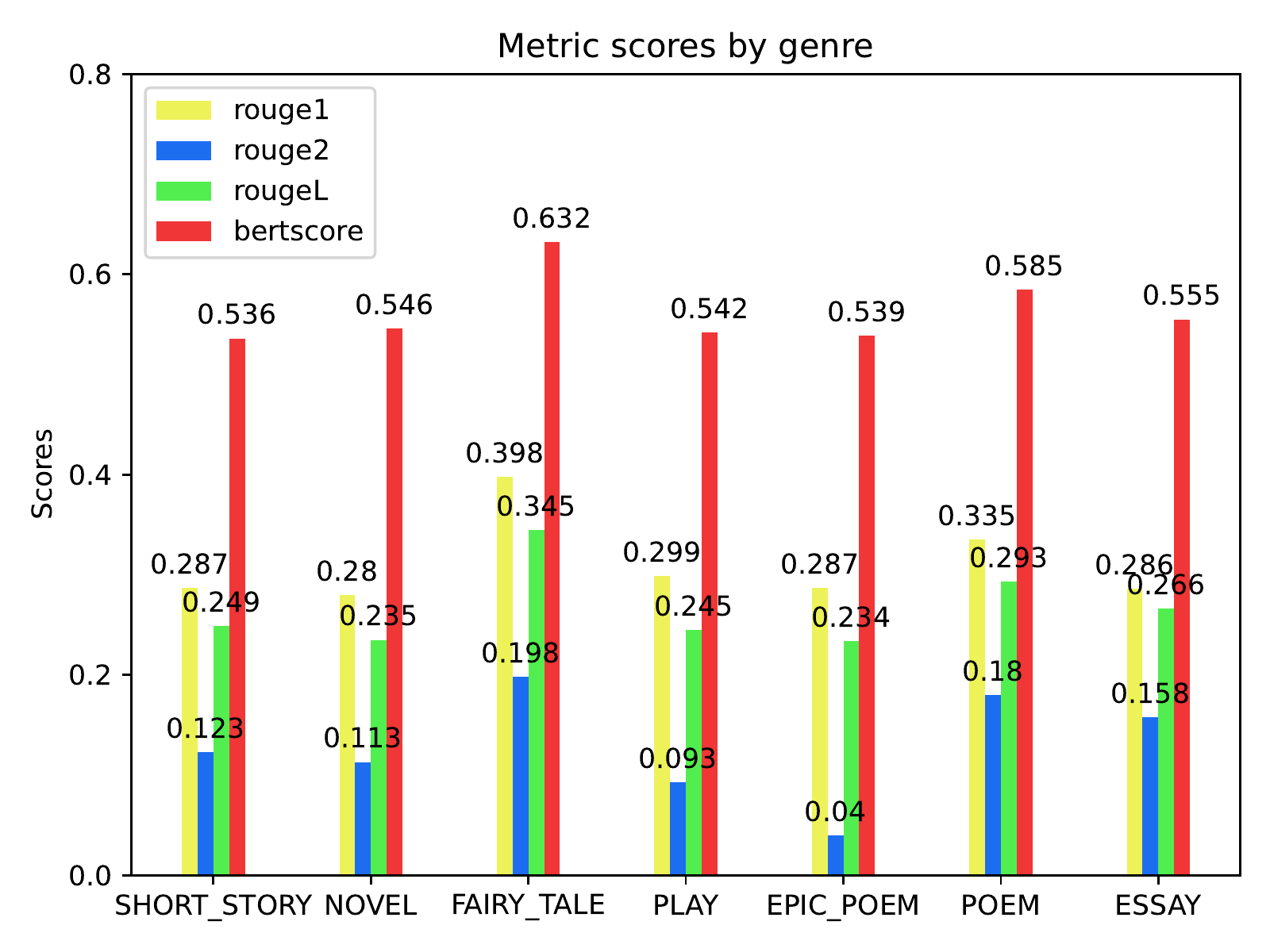}
    \caption{Genre-specific evaluation of LongT5$_{base}$ model fine-tuned on \wikiIntro. Best seen in color.}
    \vspace{-2mm}
    \label{fig:genre_specific}
\end{figure}

\paragraph{Challenges.}
\dataset{} opens the door to several other analyses and experiments that were not possible with previous datasets.
For example, we can leverage \extractive{} to perform an analysis of the behavior of the extractor submodule of our extractive-then-abstractive approach, as we show in Appendix~\ref{app:extractor_analysis}.
In Section~\ref{subsec:book-classification}, we examined the different book genres in \dataset; LongT5 model performances are detailed for each genre in
Figure~\ref{fig:genre_specific}. We notice that epic poems are the hardest to summarize in this setting, while our model performs reasonably well on fairy tales. 

\paragraph{Cross-lingual book summarization.}
Additionally, \dataset{} can be employed as a multilingual and cross-lingual summarization benchmark, thanks to its coverage of 5 languages and 25 language pairs.
In particular, we argue that cross-lingual book summarization is a very interesting challenge, as it requires a model to compress vast amounts of information while transferring knowledge across languages.
Moreover, enabling cross-lingual book summarization is fundamental for all those cases in which we do not have the source text available in the language of interest, i.e., its translation may still be under copyright or may not exist at all.
To move the first step in this direction, we propose a \textit{summarize-then-translate} approach, a simple baseline for cross-lingual book summarization on Echo-XSum.
As the name implies, our approach works by employing a monolingual model to produce a summary in the same language as the source text, and then it translates the summary from the source language to the desired target language.
We report the results of this baseline in Table~\ref{tab:summarize-then-translate}.
While this is a strong baseline, it is still affected by two main issues: i) it requires two systems, a summarizer and a translator; ii) machine translation usually fails to translate language-specific items, e.g., character names may not be exact translations.

\section{Conclusion}
In this paper, we introduced \dataset{}, the first multilingual resource for book summarization and the largest among the English datasets. 
\dataset\ features three novel datasets, namely, \wiki, \wikiIntro{}, and \extractive{}, which address several limitations of existing book summarization resources, such as BookSum.
Indeed, previous datasets for full-text book summarization are, i) limited in size, and, ii) monolingual, i.e., usually covering English only.

In addition, we leveraged \dataset{} to bring to light the unsatisfying capabilities of current approaches to generalize to book summarization. 
Finally, to mitigate this issue, we proposed a new \textit{extractive-then-abstractive} baseline for book summarization, which outperforms its purely-abstractive counterpart on \wiki{} and \wikiIntro{}, achieving results on the standard BookSum test set that are comparable with the current state of the art while using a number of parameters that is only 0.1\% compared to the best-performing method.

We believe that \dataset{} will foster future work on long-document summarization, especially in the multilingual and cross-lingual setting.


\section*{Limitations}
Despite the multilinguality of our resource, there is still a strong bias towards the English language, as the majority of books are in English and many translations are from English. This may result in the values of English literature being reflected, and these may differ from those of other cultures; summarizing literature from different cultures and regions may not be fully accurate, as every region has had its own historical development. 

Language models used in the experiments can inherit biases from the training data and the tools, such as the ones used for preprocessing, and have limitations that have not been fully evaluated and could impact the results of this study.

This study includes the use of Web data, which -- while marked as public domain -- may be subject to copyright laws.  The data used in this study was collected for research purposes and was not intended for any other use. 
Additionally, it is worth noting that the majority of books used in our resource are copyright-free, and therefore, old. While this allowed us to include a large number of texts in our dataset, it also means that our resource may not fully capture contemporary literature and may not be representative of current linguistic trends and cultural values.

\section*{Acknowledgements}
\begin{center}
\noindent
    \begin{minipage}{0.1\linewidth}
        \begin{center}
            \includegraphics[scale=0.2]{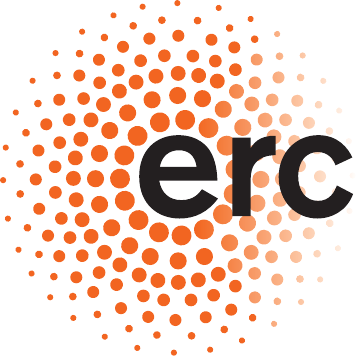}
        \end{center}
    \end{minipage}
    \hspace{0.01\linewidth}
    \begin{minipage}{0.70\linewidth}
        The authors gratefully acknowledge the support of the ERC Consolidator Grant MOUSSE No.\ 726487 under the European Union's Horizon 2020 research.
    \end{minipage}
    \hspace{0.01\linewidth}
    \begin{minipage}{0.1\linewidth}
        \begin{center}
            \includegraphics[scale=0.08]{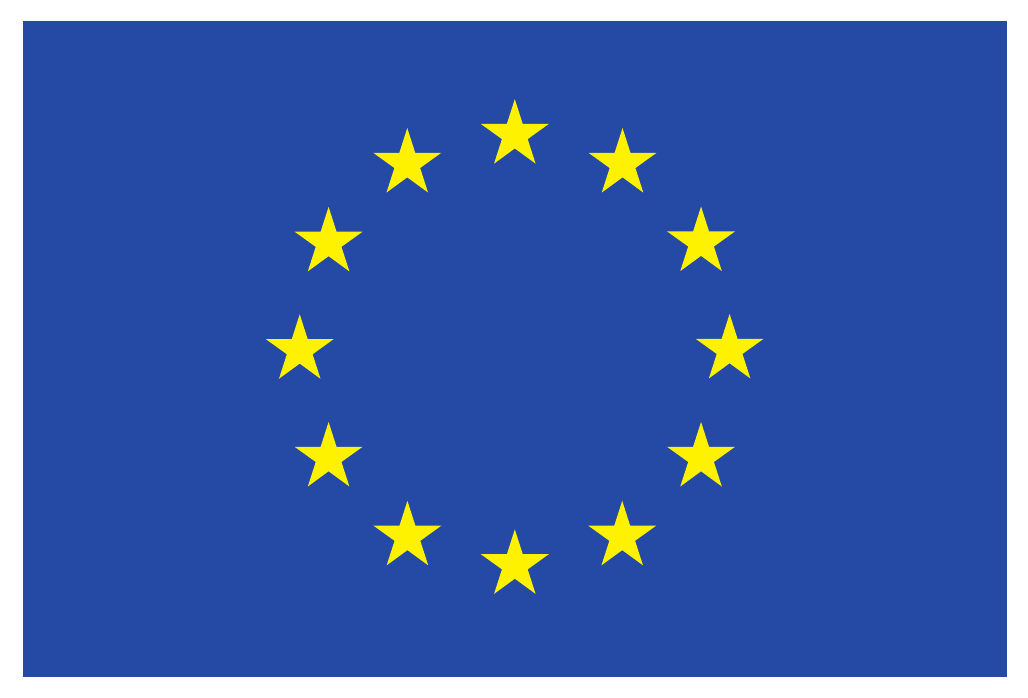}
        \end{center}
    \end{minipage}\\
\end{center}
\vspace{0.2cm}
The last author gratefully acknowledges the support of the PNRR MUR project PE0000013-FAIR. This work was carried out while Alessandro Scir\`e was enrolled in the Italian National Doctorate on Artificial Intelligence run by Sapienza University of Rome.
We would like to express our gratitude to Luigi Procopio and Edoardo Barba for their valuable insights on extractive-then-abstractive architectures, as well as to Fabrizio Brignone (Babelscape) for his exceptional support with the adaptation and use of Babelscape's keyword and phrase annotation interface.

\bibliography{main}
\bibliographystyle{acl_natbib}


\appendix
\section{Wikipedia summary sections}
\label{app:section_names}
In Table \ref{tab:section_list} we provide the list of Wikipedia section titles whose contents are used as summaries in \wiki.
\begin{table*}[ht]
\centering
\begin{adjustbox}{max width=\linewidth}
\begin{tabular}{ c c c c c }
\toprule
\textbf{IT} & \textbf{EN} & \textbf{ES} & \textbf{FR} & \textbf{DE}\\
\midrule
trame & plot overview & resumen de la trama & trame & zusammenfassung\\
trama & subject & trama & résumé synthétique & synthese \\
trama del racconto & plots & argumento & résumé & handlung\\
sinossi & plot details & contenido & trame romanesque & inhalt\\
vicenda & structure and plot & resumen & synopsis &\\
riassunto & plot and structure & sinopsis & la trame romanesque &\\
racconto & abstracts & & la trame de l’histoire &\\
il racconto & plot summary & & &\\
riassunti & synopsis & & &\\
 & subjects & & &\\
 & plot & & &\\
 & story & & &\\
 & summaries & & &\\
 & abstract & & &\\
 & the story & & &\\
 & plot synopsis & & &\\
 & plot introduction & & &\\
 & summary & & &\\
 & thematic summary & & &\\
 & summary and themes & & &\\
 & plot outline & & &\\
\bottomrule
\end{tabular}
\end{adjustbox}
\caption{Table of Wikipedia section titles utilized in the \wiki~parsing process in multiple languages}
\label{tab:section_list}
\end{table*}
\begin{figure}[t]
    \centering
    \includegraphics[width=1.0\linewidth]{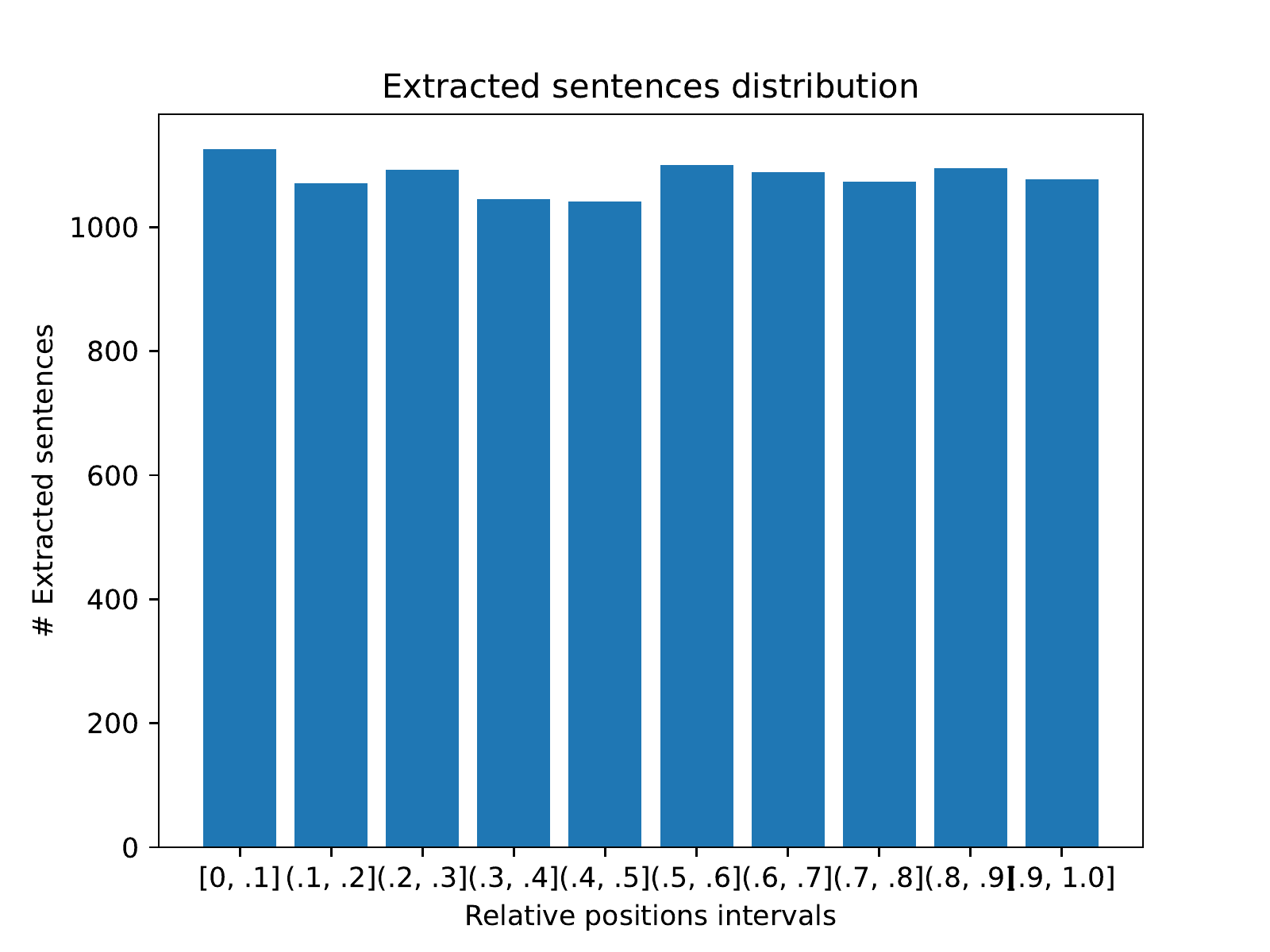}
    \caption{Number of extracted sentences for each relative position interval.}
    \label{fig:extractor_distribution}
\end{figure}
\section{\wikiIntro{} example}
\begin{figure*}[h]
    \centering
    \includegraphics[scale=0.8]{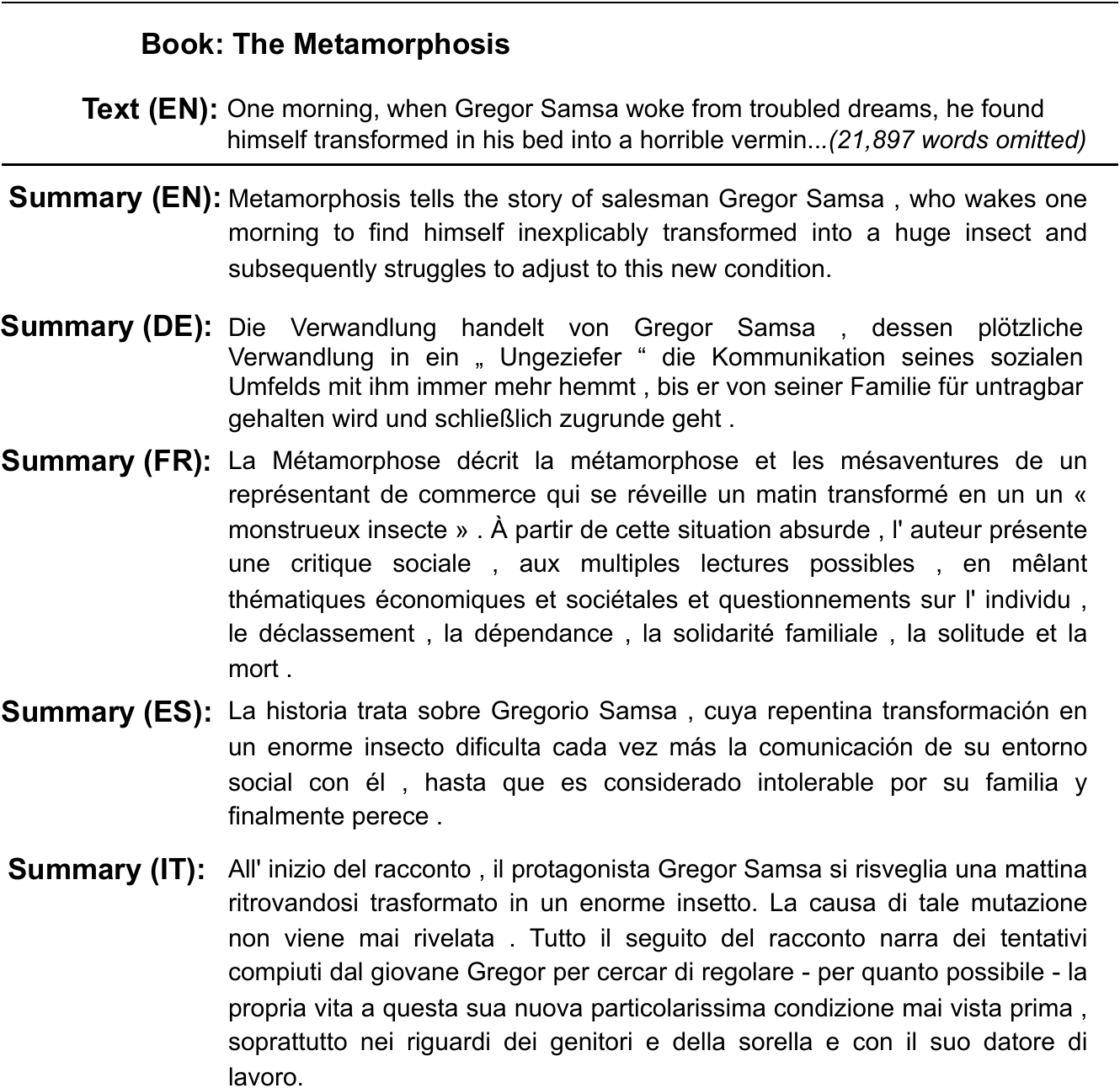}
    \caption{An excerpt of a book text along with multilingual summaries from \wikiIntro{}.}
    \label{fig:echoxsum_extract}
\end{figure*}

In Figure \ref{fig:echoxsum_extract} we report an excerpt of the book text of the English version of "The Metamorphosis" by Franz Kafka, along with the multilingual extreme summaries from \wikiIntro{}.
\label{app:echoxsum_excerpt}

\section{\wikiIntro{}~annotation task}
\label{app:echo_xsum_annotation}
\begin{figure*}[h]
    \centering
    \includegraphics[scale=0.4]{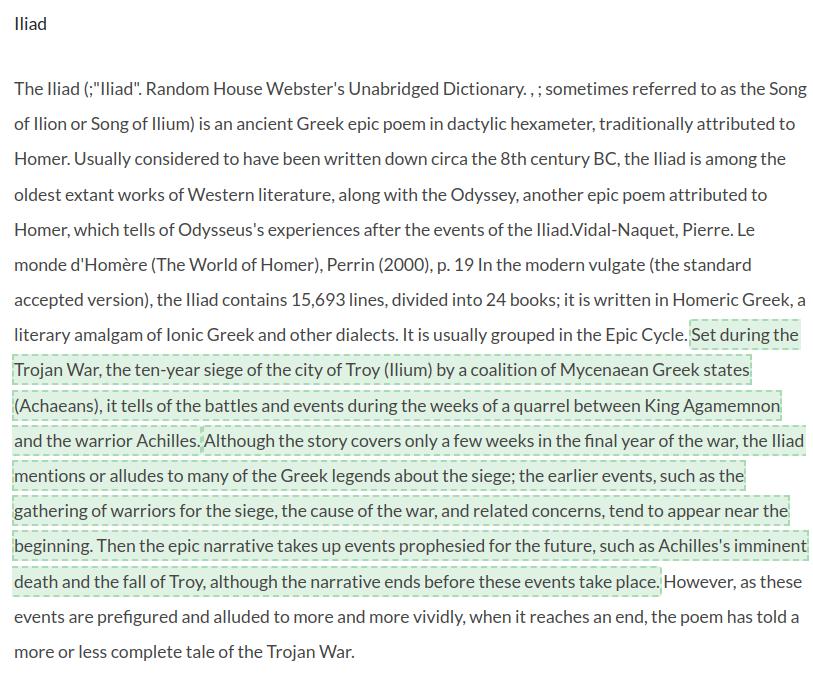}
    \caption{\wikiIntro~annotation process consists of highlighting plot-specific pieces of text from the lead section of the Wikipedia page.}
    \label{fig:echoxsum_annotation}
\end{figure*}

In Figure \ref{fig:echoxsum_annotation} we provide an example of a manually-annotated summary in \wikiIntro{}. The annotator was tasked to highlight portions of text containing information related to the plot from the Wikipedia introduction.

\section{Extractor analysis}
\label{app:extractor_analysis}
We analyze the positions of the sentences selected by the extractor. This analysis is required to investigate the presence of any positional bias, e.g., the lead bias, which is known to affect systems trained on news stories. Figure \ref{fig:extractor_distribution} depicts the distribution of the relative positions of the extracted sentences on texts from \extractive{}, i.e., fairy tales and short stories. We deduce that the extractions are not affected by any bias.
Thanks to \extractive{} extractive annotations, we are also able to evaluate the performance of the extractor component of the \textit{extractive-then-abstractive} approaches.
We aggregate multiple extractive annotations in \extractive{} by retaining the intersecting sentences; we refer to these sentences as the gold sentences. We measure the Extractor performance by computing the overlap between the sentences extracted by the model and the gold ones. We compute the \textit{Precision@K} by comparing the topK-ranked sentences with the references. We report the Extractor performance in Table~\ref{tab:extractor_evaluation}. We observe relatively low scores, meaning that the extractor is only partially able to discriminate relevant sentences from irrelevant ones. This aspect confirms that there is still large room for improving the Extractor and, consequently, the relevance of the summaries. 

\begin{table}[h!]
\centering
\begin{adjustbox}{max width=\linewidth}
\begin{tabular}{ c c c }
\toprule
\textbf{K} & \textbf{Precision}\\
\midrule
1 &  31.1\\ 
2 &  28.8\\ 
3 &  28.8\\ 
4 &  27.2\\ 
5 &  25.6\\ 
\bottomrule
\end{tabular}
\end{adjustbox}
\caption{Extractor evaluation: Precision@K}
\label{tab:extractor_evaluation}
\end{table}

\end{document}